%% file: sample-sigconf.tex
\documentclass[sigconf]{acmart}

\usepackage{booktabs} 

\setcopyright{none}

\newcommand{\Sect}[1]{Section~\ref{#1}}
\renewcommand{\paragraph}[1]{\textbf{#1}~~}

\newcommand{\Fig}[1]{Fig.~\ref{#1}}

\begin{document}
\title{Mobile Machine Learning Hardware at ARM:
\\A Systems-on-Chip (SoC) Perspective}

\author{Yuhao Zhu}
\authornote{Work done while a visiting researcher at ARM Research}
\affiliation{%
  \department{Department of Computer Science}
  \institution{University of Rochester}
}
\email{yzhu@rochester.edu}

\author{Matthew Mattina}
\affiliation{%
  \department{Machine Learning \& AI}
  \institution{ARM Research}
}
\email{matthew.mattina@arm.com}

\author{Paul Whatmough}
\affiliation{%
  \department{Machine Learning \& AI}
  \institution{ARM Research}
}
\email{paul.whatmough@arm.com}

\begin{abstract}
Machine learning is playing an increasingly significant role in emerging mobile application domains such as AR/VR, ADAS, etc. Accordingly, hardware architects have designed customized hardware for machine learning algorithms, especially neural networks, to improve compute efficiency. However, machine learning is typically just one processing stage in complex end-to-end applications, involving  multiple components in a mobile Systems-on-a-chip (SoC). Focusing only on ML accelerators loses bigger optimization opportunity at the system (SoC) level. This paper argues that hardware architects should expand the optimization scope to the entire SoC. We demonstrate one particular case-study in the domain of \textit{continuous computer vision} where camera sensor, image signal processor (ISP), memory, and NN accelerator are synergistically co-designed to achieve optimal system-level efficiency. 
\end{abstract}

\maketitle

\input{samplebody-conf}

\bibliographystyle{ACM-Reference-Format}
\bibliography{sample-bibliography} 

\end{document}

%% file: samplebody-conf.tex

\section{Introduction}
\label{sec:intro}

Mobile devices are the most prevalent computing platform of the present day, and are dominated by the ARM architecture.
A large number of emerging mobile application domains now rely heavily on machine learning; in particular, various forms of deep neural networks (DNNs) have been instrumental in driving progress on problems such as computer vision and natural language processing.
On mobile platforms, DNN inference is currently typically executed in the cloud. However, the trend is to move DNN execution from the cloud to the mobile devices themselves. This shift is essential to remove the communication latency and privacy issues of the cloud offloading approach.

The increasing use of DNNs in mobile applications places significant compute requirements on the mobile System-on-chip (SoC), which must process tens of billions of linear algebra operations per second within a tight energy budget.
In response, there has been significant effort expended on dedicated hardware to accelerate the computation of neural networks. This is borne out in a proliferation of designs for DNN accelerators (NNX), which typically demonstrate high computational efficiency on the order of 0.4 -- 3.8 TOPS/W on convolutional NN inference \cite{eyerisschip, whatmough, stmdcnn, envision}. This is several orders of magnitude more efficient than typical mobile CPU implementations. 

Sadly, the efficiency benefits of hardware accelerators are largely a one-time improvement, and will likely saturate, while the compute requirement of DNNs keep increasing. 
Using computer vision as an example, today's convolutional neural network (CNN) accelerators are not able to perform object detection (e.g., YOLO\cite{yolo}) in real time at 1080p/60fps. 
As the resolution, frame rate, and the need for stereoscopic vision grows with the emergence of AR/VR use cases, the compute requirement will continue to increase, while the power budget remains constant, leaving a large gap.


Therefore, we must move from a narrow focus on hardware accelerators to begin to consider system-level optimizations for ML on mobile.
Expanding the scope beyond the DNN accelerator to consider the whole SoC, we emphasize three areas for optimization:
\begin{itemize}
  \item \paragraph{Accelerator Interfacing}: hardware accelerators must be efficiently interfaced with the rest of the SoC for full benefit.
  \item \paragraph{Software Abstractions}: for cross-platform compatability, SoC details should be abstracted with a clean API.
  \item \paragraph{System Optimizations}: Co-design of algorithms and the various hardware blocks in the system. 
\end{itemize} 
\Sect{sec:soc} will describe these optimizations further, with a case study in \Sect{sec:case}. We conclude in~\Sect{sec:conc}.

\section{ML on Mobile Systems}
\label{sec:soc}

\begin{figure}[t]
  \centering
  \includegraphics[trim=0 0 0 0, clip, width=\columnwidth]{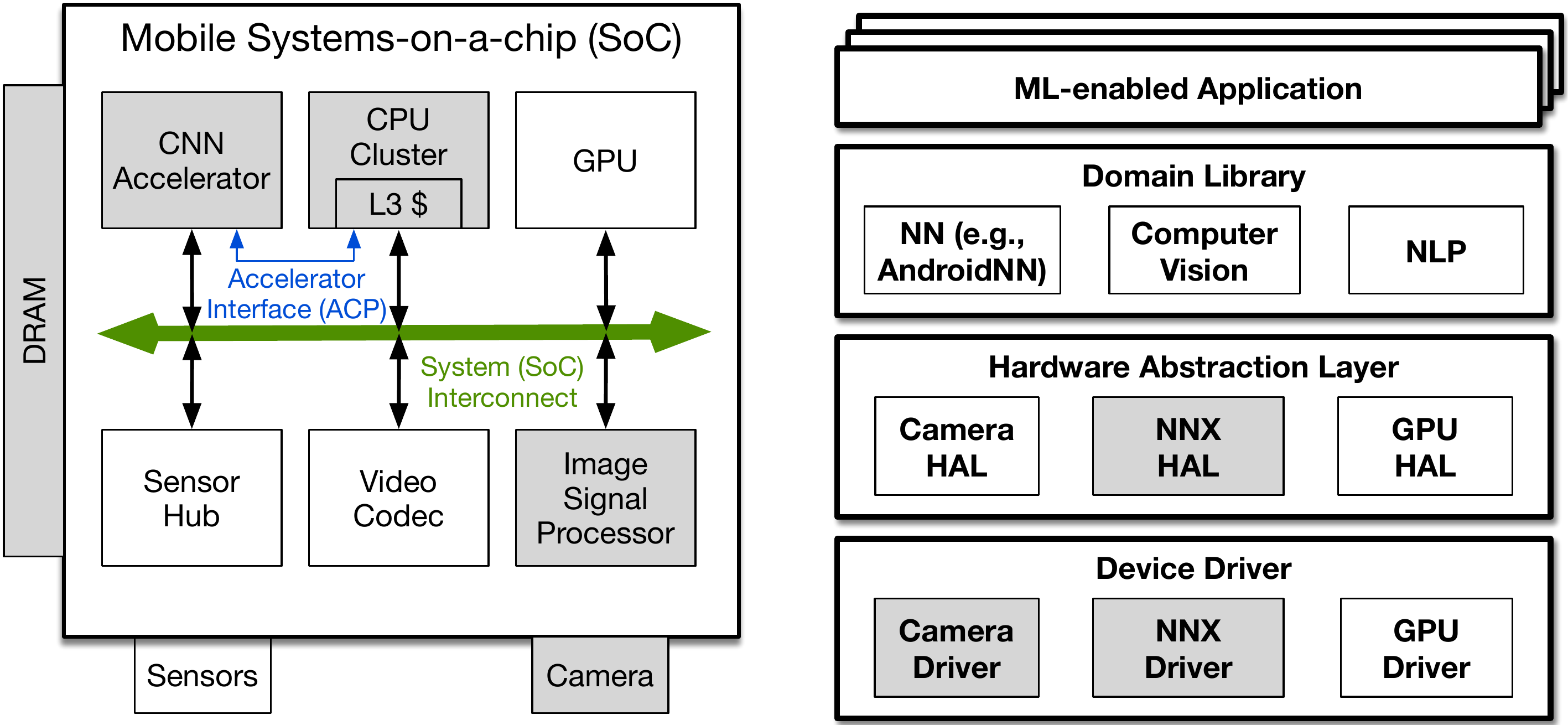}
  \caption{Mobile SoC (left) and software stack (right). Shaded components are used in continuous vision tasks.}
  \label{fig:soc}
\vspace{-10pt}
\end{figure}

We have already started to see changes in mobile systems in response to the computational demands of deep learning methods. Most notably, NNX components are now common in mobile SoCs, e.g., the Neural Engine in the iPhoneX~\cite{applexaicore} and the HPU CNN co-processor in the Microsoft HoloLens~\cite{hpuaicore}. However, a significant challenge still remains as to how to integrate NNX components into the system. We identify three aspects from both hardware and software perspectives.


\paragraph{Accelerator Interfacing} There are two main challenges in interfacing NNX IPs to the rest of the mobile hardware system:
(1) providing an efficient path to offload an NN invocation task from the CPU to the NNX, and (2) providing sufficient memory bandwidth for weights and activation data. 
In particular, the cache/memory interface between the accelerator and the SoC is critical, since modern DNNs typically have very large parameter sets, which demand high memory bandwidth to feed many arithmetic units~\cite{diannao}.

Our key insight is to leverage the L3 cache in the CPU cluster as a bandwidth filter rather than directly interfacing the NNX with the DRAM as some state-of-the-art NNX designs currently do~\cite{eyeriss}. 
This is achieved through the ARM Accelerator Coherency Port (ACP)~\cite{acp}, which is available on ARM CPU clusters and allows attached accelerators to load and store data directly to and from the L3. 
In this way, the NNX can also take advantage of L3 cache features in the CPU cluster such as prefetching,cache stashing, and partitioning.
Furthermore, ACP is also low-latency, such that the CPUs and NNX can work together closely on data in the L3 cache.

\paragraph{Software Abstractions} Mobile SoC hardware and ML algorithms are both evolving rapidly. 
Therefore, it is paramount to present a programming interface that minimizes disruption to application developers. This would make new hardware features easy to use, and provide compatibility across a range of mobile SoCs.

The key of such a programming interface is a clear abstraction that allows applications to execute DNN jobs efficiently on (one of many) hardware accelerators, or fall back to execution on a CPU or GPU.
The AndroidNN API~\cite{androidnn} provides an example of this principle, by abstracting common DNN kernels such as convolution, and scheduling execution through a hardware abstraction layer (HAL).

We provide optimized implementations for DNN kernels in the form of ARM Compute Library~\cite{armcl}. The library takes advantage of recent ARM ISA enhancements that provides new instructions for essential linear algebra operations behind DNNs, such as the new dot-product instructions in Arm's NEON SIMD extension~\cite{armneon}. Finally, we provide IP-specific drivers and HAL implementations to support the AndroidNN API.


\paragraph{System Optimizations} While adding specialized NNX hardware IP to the SoC improves kernel-level performance and efficiency, the DNN component is typically only one stage in a larger end-to-end application pipeline. 
For instance in computer vision, many on/off-chip components such camera sensors, Image Signal Processors (ISP), DRAM memory, as well as the NNX have to collaborate together to deliver real time vision capabilities. 
The NNX IP itself constitutes at most half of the total power/energy consumption, and there are additional opportunities in jointly optimizing the whole system.

Once we expand our scope to the system level, we expose new optimization opportunities by exploiting functional synergies across different IP blocks; these optimizations are not obvious when considering the NNX in isolation.
We will demonstrate this principle in the following case study.

\section{Case Study: Continuous Vision}
\label{sec:case}

Computer vision (CV) tasks such as object classification, localization, and tracking are key capabilities for a number of exciting new application domains on mobile devices, such as augmented reality (AR) and virtual reality (VR).
However, the computational cost of modern CV CNNs far exceed the severely limited power budget of mobile devices, especially for real time (e.g., 1080p/60fps). 
This is true even with a dedicated CNN hardware accelerator~\cite{eyerisschip, stmdcnn, envision}.



To achieve real-time object detection with high accuracy on mobile devices, our key idea is to reduce the total number of expensive CNN inferences through system-level optimizations. 
This is done by harnessing the synergy between different hardware IPs of the vision subsystem.
In particular, we leverage the fact that the image signal processor (ISP) inherently calculates motion vectors (MV) for use in its temporal denoising algorithms.
Usually MVs are discarded after de-noising, but we elect to expose them at the system level.
Instead of using CNN inference on every frame to track the movement of objects, we reuse the MVs to extrapolate the movement of objects detected in the previous video frame, without further CNN inferencing for the current frame.
As we increase the number of consecutively extrapolated frames (extrapolation window, or EW), the total number of CNN inferences is reduced, leading to performance and energy improvements.

We also leverage the ARM ACP interface~\cite{acp} to use the LLC for inter-layer data reuse (e.g., feature maps), which would otherwise be spilled to the DRAM from the NN accelerator's local SRAM. A typical L3 size in mobile devices is about 2~MB~\cite{mobilecpu} and ACP provides around 20 GB/s of bandwidth, which is sufficient to capture the reuse of most layers in today's object detection CNNs. This design greatly reduces DRAM and system power consumption.

Finally, we present software support that abstracts away the hardware implementation details. As~\Fig{fig:soc} shows, the high-level CV libraries are unmodified, keeping the AndroidNN interface unchanged. We implement specific driver and HAL modifications that our hardware augmentation entails.

\begin{figure}[t]
  \centering
  \includegraphics[trim=0 0 0 0, clip, width=.9\columnwidth]{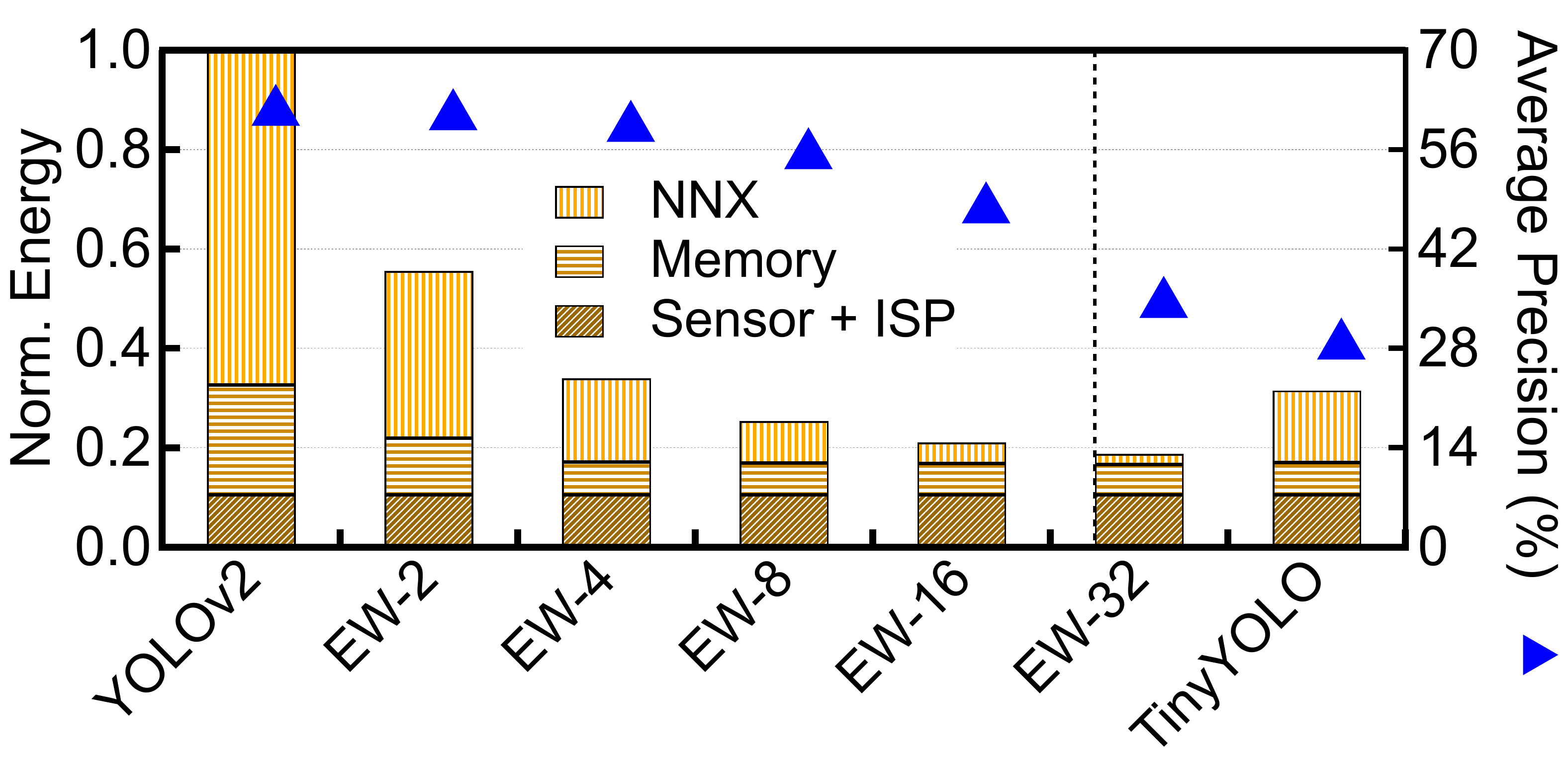}
  \caption{Cross-IP optimization of object detection on a mobile SoC allows over 40\% reduction in energy (left) with less than 1\% accuracy loss (right).}
  \label{fig:res}
\vspace{-20pt}
\end{figure}

We evaluated the system-level optimizations on an in-house SoC simulator, which we calibrated with measurements on the Jetson TX2 development board~\cite{tx2}. We use commonly-used benchmarks such as VOT~\cite{vot2014benchmark} and OTB~\cite{otbdataset} as well as our internal datasets.
Results in~\Fig{fig:res} show that compared to state-of-the-art
object detection frameworks such as YOLO~\cite{yolo} that execute an entire CNN for every frame, our system reduces the energy by over 40\% with less than 1\% accuracy loss at an extrapolation window (EW) size of two. 
The energy saving is greater as EW increases, while accuracy degrades. 
Compared to the conventional approach of reducing the compute intensity by down-scaling the network (e.g., TinyYOLO, which is $\sim$ 5 $\times$ simpler), our system achieves higher energy savings and higher accuracy.

\section{Conclusion}
\label{sec:conc}

Efficiently supporting demanding ML workloads on energy-constrained mobile devices requires careful attention to the overall system design. 
We emphasized three key research priorities: accelerator interfacing, software abstractions, and cross-IP optimizations.
